\documentclass[12pt]{article}

\usepackage{times}
\usepackage{graphicx} 
\usepackage{subfigure} 
\usepackage{multirow}
\usepackage[nomarkers,nofiglist,notablist]{endfloat}
\usepackage{natbib}

\usepackage{algorithm}
\usepackage{algorithmic}
\usepackage[breaklinks=true]{hyperref}

\usepackage{bbm}
\usepackage{mathtools}
\usepackage{amsmath}
\usepackage{amsfonts}
\usepackage[most]{tcolorbox}
\usepackage{nicefrac}
\usepackage{bm}

\def\abovestrut#1{\rule[0in]{0in}{#1}\ignorespaces}
\def\belowstrut#1{\rule[-#1]{0in}{#1}\ignorespaces}

\def\abovespace{\abovestrut{0.20in}}

\def\belowspace{\belowstrut{0.10in}}

\renewcommand{\cite}[1]{\citep{#1}}

\graphicspath{{./figures/}{.}}

\usepackage[margin=.93in]{geometry}
\usepackage{authblk}

\title{More than one Author with different Affiliations}
\author[1]{Tammo Rukat\thanks{tammo.rukat@stats.ox.ac.uk}}
\author[1]{Chris C. Holmes\thanks{cholmes@stats.ox.ac.uk}}
\author[2]{Michalis K. Titsias\thanks{mtitsias@aueb.gr}}
\author[3,4]{Christopher Yau\thanks{c.yau@bham.ac.uk}}
\affil[1]{Department of Statistics, University of Oxford, UK}
\affil[2]{Department of Informatics, Athens University of Economics and Business, Greece}
\affil[3]{Centre for Computational Biology, Institute of Cancer and Genomic Sciences, University of Birmingham, UK}
\affil[4]{Wellcome Trust Centre for Human Genetics, University of Oxford, UK}

\date{\vspace{-5ex}}

\newcommand*\wc{{\mkern 2mu\cdot\mkern 2mu}}
\begin{document}

\title{Bayesian Boolean Matrix Factorisation}
\maketitle

\begin{abstract}
  Boolean matrix factorisation aims to decompose a binary data matrix into an approximate Boolean product of two low rank, binary matrices: one containing meaningful patterns, the other quantifying how the observations can be expressed as a combination of these patterns.
  We introduce the OrMachine, a probabilistic generative model for Boolean matrix factorisation and derive a Metropolised Gibbs sampler that facilitates efficient parallel posterior inference. On real world and simulated data, our method outperforms all currently existing approaches for Boolean matrix factorisation and completion. This is the first method to provide full posterior inference for Boolean Matrix factorisation which is relevant in applications, e.g.\ for controlling false positive rates in collaborative filtering and, crucially, improves the interpretability of the inferred patterns. The proposed algorithm scales to large datasets as we demonstrate by analysing single cell gene expression data in 1.3 million mouse brain cells across 11 thousand genes on commodity hardware.
\end{abstract}

\section{Introduction}

%

\begin{figure}[tb]
  \centering
  \begin{minipage}{.93\linewidth}
    \centering
    \includegraphics[width=\linewidth]{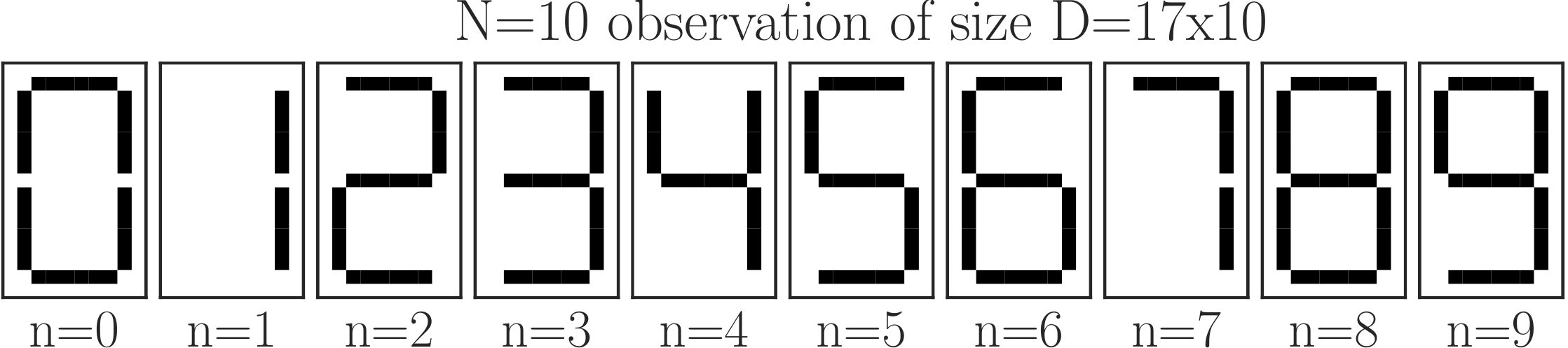}
    \vspace{.01mm}
  \end{minipage}
  
  \begin{minipage}{.02\linewidth}
    $\simeq$ \vspace{.2cm}
  \end{minipage}\hspace{.03\linewidth}
  \begin{minipage}{.3\linewidth}
    \centering
    \vspace{-.2\linewidth}
    \includegraphics[width=\linewidth]{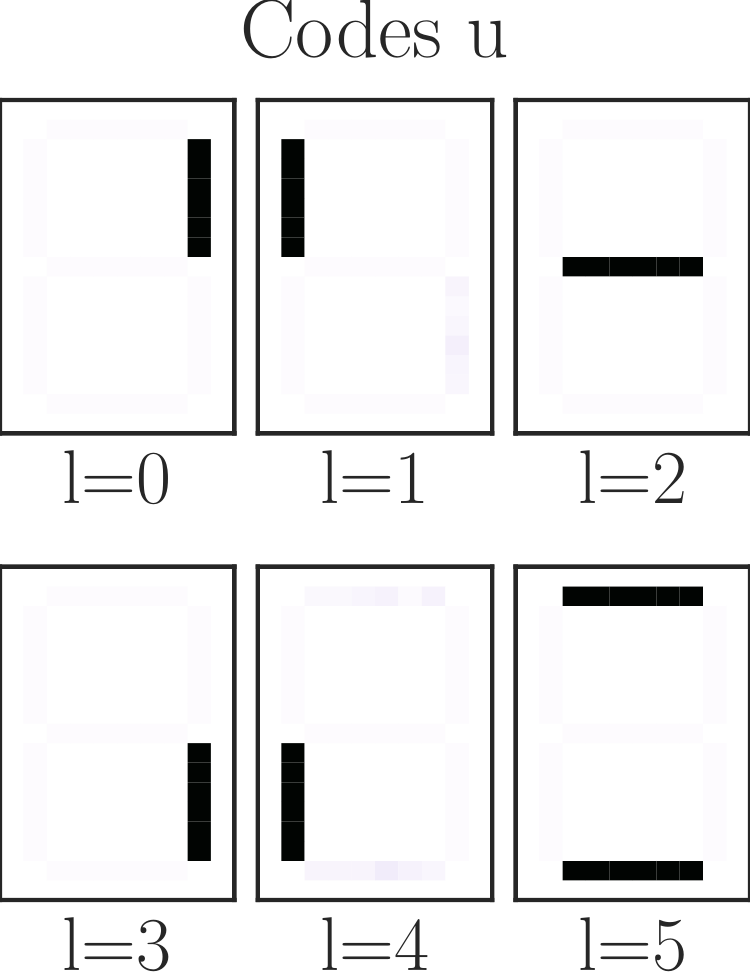}    
  \end{minipage}\hspace{.1cm}
  \begin{minipage}{.02\linewidth}
    $\otimes$  \vspace{.4cm}
  \end{minipage}\hspace{.2cm}
  \begin{minipage}{.5\linewidth}
    \centering
    \vspace{-.06\linewidth}
    \includegraphics[width=\linewidth]{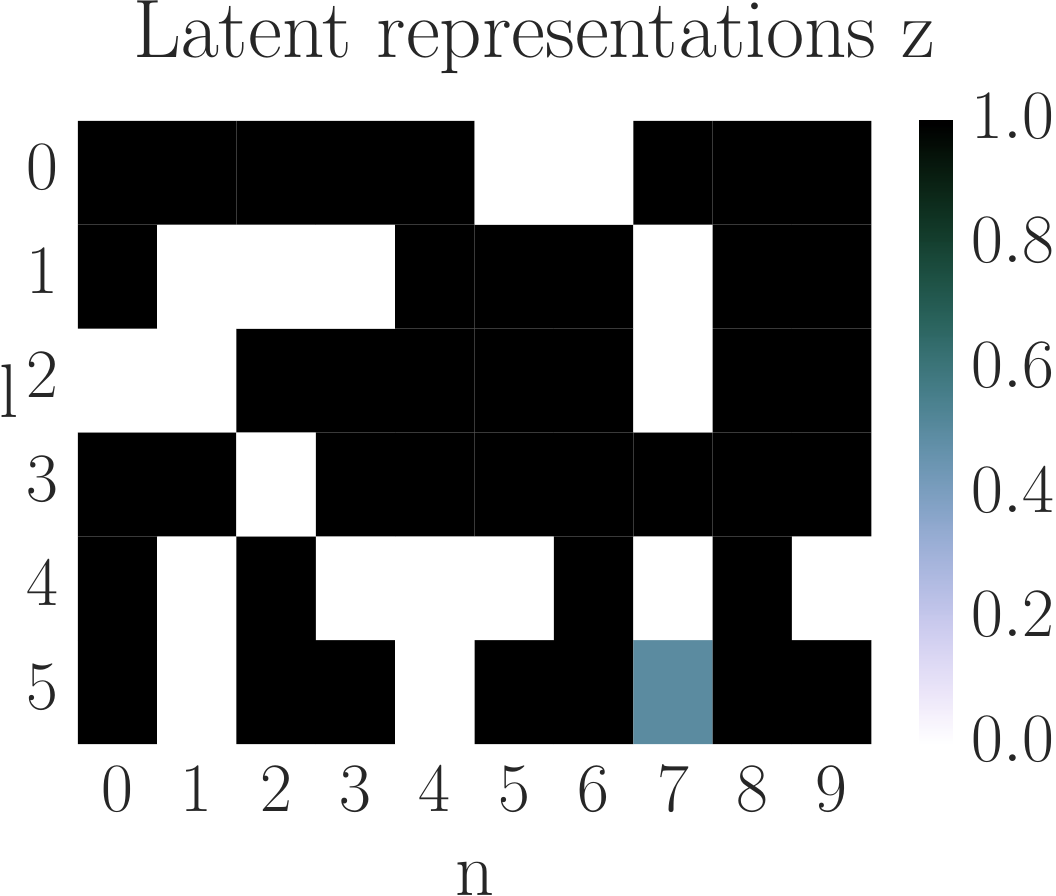}    
  \end{minipage}
  \caption{The observed images are 10 digits from $0$ to $9$ as they are traditionally represented in calculators. The data is factorised into matrices of rank 6, which is not sufficient for full error-free reconstruction. Every digit, except $7$, can be constructed by Boolean combination of the inferred codes. The OrMachine infers a posterior mean probability of $50\%$ for using code $l=5$ in constructing a $7$. Note that there exist other equally valid solutions to this problem with 6 latent dimensions.  The pixels represent posterior means. Codes and observations are arranged to $10{\times}17$ images for interpretation. 
 \label{fig:calc}}
\end{figure}




Boolean matrix factorisation (BooMF) can infer interpretable decompositions of a binary data matrix \mbox{$\bm{X}\in\{0,1\}^{N\times D}$} into a pair of low-rank, binary matrices \mbox{$\bm{Z}\in\{0,1\}^{N\times L}$} and \mbox{$\bm{U}\in\{0,1\}^{D\times L}$}. The data generating process is based on the Boolean product, a special case of matrix product between binary matrices where all values larger than zero are set to one, i.e.\
\begin{align}
  x_{nd} = \bigvee\limits_{l=1}^L z_{nl} \land u_{ld}\;. \label{eq:bool}
\end{align}

Here, $\lor$ and $\land$ encode the Boolean disjunction and conjunction, respectively. 
BooMF provides a framework for learning from binary data 
where the inferred codes $\bm{U}$ provide a basis and the indicator variables $\bm{Z}$ encode the presence or absence of these codes. This representation is illustrated in the calculator digits example in Fig.~\ref{fig:calc}. We can think of BooMF as binary factor analysis or as clustering with joint assignments, where each observation is assigned to a subset of $L$ cluster centroids or codes. The $L$-dimensional indicators provide a compact representation of which codes are allocated to each observation. As stated in Eq.~\eqref{eq:bool}, a feature $x_{nd}$ takes a value of one if it equals one in any of the assigned codes.

BooMF has many real-world applications ranging from topic modelling~\cite{blei2012_probab} to collaborating filtering~\cite{su2009_survey-collab} and computer vision \cite{lazaro-gredilla2016_hierar}.
In this paper, we introduce the OrMachine, a Bayesian approach to BooMF, and fit the model using a fast and scalable Metropolised Gibbs sampling algorithm. On simulated and real-world data, our method is shown to significantly outperform the current state-of-the-art message passing approaches for learning BooMF models.
Moreover, we consider a challenging application in the analysis of high-throughput single cell genomics data. BooMF is used to identify latent gene signatures (codes) that correspond to key cellular pathways or biological processes from large gene expression datasets consisting of 1.3 million cells across 11 thousand genes. Genes are expressed if one or more relevant biological processes are active, a property which is naturally modelled by the Boolean OR operation. We also introduce a multi-layered extensions of Bayesian BooMF that can capture hierarchical dependencies in the latent representations.

\section{Related Work}
\label{sec:related_work}

There has been a sustained interest in BooMF and related methods of which we will give a brief review.
The Discrete Basis Problem \cite{miettinen2006_discr-basis-probl} provides a greedy heuristic algorithm to solve BooMF without recourse to an underlying probabilistic model. It is based on association rule mining \cite{agrawal1994fast} and has more recently been extended to automatically select the optimal dimensionality of the latent space based on the minimum description length principle \cite{miettinen2014_mdl4b}. In contrast, multi assignment clustering for Boolean data \cite{streich09:_multi_boolean} leverages on a probabilistic model for BooMF, adding a further global noise source to the generative process. Point estimates are inferred by deterministic annealing. Similarly, \citet{wood2012_non-param} develop a probabilistic model to infer hidden causes. In contrast to the Boolean OR, the likelihood of an observation increases with the number of active hidden codes. They use an Indian Buffet process prior over the latent space and a Gibbs sampler to infer the distribution over the unbounded number of hidden causes.
A similar approach to ours is the work by \citet{ravanbakhsh2015_boolean-matrix}. The authors tackle BooMF using a probabilistic graphical model and derive a message passing algorithm to perform MAP inference. Their method is shown to have state-of-the-art performance for BooMF and completion. It therefore serves us as baseline benchmark in these tasks. The message passing approach has recently been employed by \citet{lazaro-gredilla2016_hierar} in a hierarchical network combined with pooling layers to infer the building blocks of binary images.

\section{The OrMachine} \label{sec:ormachine}
\subsection{Model Formulation}

The OrMachine is a probabilistic generative model for Boolean matrix factorisation.
A matrix of $N$ binary observations $\bm{x}_n\in\{0,1\}^D$ is generated from a discrete mixture of $L$ binary codes $\mathbf{u}_l \in \{0,1\}^{D}$. 
Binary latent variables $z_{nl}$ denote whether or not code $l$ is used in generating a particular observation $\bm{x}_n$.
The probability for a data point $x_{nd}$  to be one is greater than $\nicefrac{1}{2}$ if the corresponding codes and latent variables in at least one latent dimension both equal one;
conversely, if there exists no dimension where codes and latent variables both equal one, the probability for the data point to be one is less than $\nicefrac{1}{2}$.
The exact magnitude of this probability is inferred from the data and, for later notational convenience, is parametrised as the logistic sigmoid of a global dispersion parameter $\sigma(\lambda)=(1+e^{-\lambda})^{-1}$, with $\lambda \in \mathbb{R}^+$. Next, we give a full description of the likelihood and prior distributions used in the OrM.

The likelihood function is factorised across the $N$ observations and $D$ features 
with each factor given by 
\begin{align}
  p(x_{nd}|\bm{u},\bm{z},\lambda) &=	 
  \begin{cases}
    \sigma(\lambda);&\text{if}\;x{=}\min(1,\bm{u}_d^T\bm{z}_n)\\
    1{-}\sigma(\lambda);&\text{if}\;x{\neq}\min(1,\bm{u}_d^T\bm{z}_n)
  \end{cases}  \label{eq:lik_eff1}  \\
   &= \sigma\left[\lambda \tilde{x}_{nd} \left(1-2\prod\limits_{l}\left(1-z_{nl}u_{ld}\right)\right) \right]\;. \label{eq:lik_eff2} 
\end{align}
Tilde denotes the $\{0,1 \} \rightarrow \{-1,1\}$ mapping so that for any binary variable $x \in \{0,1\}$, 
$\tilde{x}=2x-1$.
The expression inside the parentheses of Eq.~\eqref{eq:lik_eff2} encodes the OR operation and evaluates to $1$ if $z_{nl}=u_{ld}=1$ for at least one $l$, and to $-1$ otherwise.
The dispersion parameter controls the noise in the generative process, i.e.\ as $\lambda \rightarrow \infty$, all probabilities tend to 0 or 1 and the model describes a deterministic Boolean matrix product.
Note that the likelihood can be computed efficiently from Eq.~\eqref{eq:lik_eff2} as we describe in detail 
in the next section. 
We further assume independent Bernoulli priors for all variables $u_{ld}$ and $z_{nl}$. Such priors allow us to promote denseness or sparsity in codes and latent variables. Notice that the designation of $\bm{U}$ as codes and $\bm{Z}$ as latent variables is not necessary 
since 
these matrices  
appear in a symmetric manner. 
If we transpose the matrix of observations $\bm{X}$, then codes and latent variables merely swap roles.

Finally, we do not place a prior on the dispersion parameter $\lambda$, but maximise it using an EM-type algorithm described below.

\subsection{Fast Posterior Inference}

The full joint distribution of all data and random variables is given by
\begin{align}
p(\bm{X},\bm{U},\bm{Z}|\lambda) = p(\bm{X}|\bm{U},\bm{Z},\lambda)p(\bm{U})p(\bm{Z})\;. 
\end{align}
The full conditional for $z_{nl}$ (and analogous for $u_{ld}$) is
\begin{align}
    p(z_{nl}|\wc) = \sigma \left[\lambda \tilde{z}_{nl} \sum\limits_d \tilde{x}_{nd} u_{ld}\prod\limits_{l'\neq l} (1{-}z_{nl'}u_{l'd})  + \text{logit}(p(z_{nl})) \right]\;. \label{eq:update_z}
\end{align}

Notice that the independent Bernoulli prior enters the expression as additive term inside the sigmoid function that vanishes for the uninformative Bernoulli prior $p(z)=\nicefrac{1}{2}$. 

The form of Eq.~\eqref{eq:update_z} allows for computationally efficient evaluation of the conditionals. 
The underlying principle is that once certain conditions are met, the result of the full conditional is known without considering the remainder of a variable's Markov blanket.
For instance, when computing updates for $z_{nl}$, terms in the sum over $d$ necessarily evaluate to zero if one of the following conditions is met: (i)~$u_{ld} = 0$ or (ii)~$z_{nl'}u_{l'd} = 1$ for  some $l' \neq l$. 
This leads to Algorithm \ref{alg:update_z_2} for fast evaluation of the conditionals.
\begin{algorithm}
  \caption{Computation of the full conditional of $z_{nl}$}
  \label{alg:update_z_2}
\begin{algorithmic}[tb]
  \STATE{$\text{accumulator}=0$}
  \FOR{$d\;\text{in}\;1,\ldots, D$}
  \IF{$u_{ld} = 0$}
  \STATE continue (next iteration over $d$)
  \ENDIF
  \FOR{$l'\;\text{in}\;1,\ldots, L$}
  \IF{$l' \neq l\;\text{and}\;z_{nl'}=1\;\text{and}\;u_{l'd}=1$}
  \STATE continue (next iteration over $d$)
  \ENDIF
  \ENDFOR
  \STATE $\text{accumulator} =  \text{accumulator} + \tilde{x}_{nd} $
  \ENDFOR \STATE{$p(z_{nl}|\wc) = \sigma\left(\lambda\cdot\tilde{z}_{nl}\cdot\text{accumulator} \right)$}
\end{algorithmic}
\end{algorithm}

To infer the posterior distribution over all variables $u_{ld}$ and $z_{nl}$ we could iteratively sample from the above conditionals using standard Gibbs sampling. In practice we use a modification of this procedure which is referred to as Metropolised Gibbs sampler and was proposed by~\citet{liu96:_miscel}.
We always propose to flip the current state, leading to a Hastings acceptance probability of  $p(z|\wc)/(1-p(z|\wc))$. This is guaranteed to yield lower variance Monte Carlo estimates~\cite{peskun1973optimum}.

After every sweep through all variables, the dispersion parameter $\lambda$ is updated to maximise the 
likelihood akin to the M-step of a Monte Carlo EM algorithm. Specifically, given the current values of the
codes $\bm{U}$ and latent variables $\bm{Z}$ we can compute how many observations $x_{nd}$ are correctly predicted by the model, as
\begin{align}
 P = \sum\limits_{n,d} \mathbbm{I}\left[x_{nd}=1-\prod\limits_{l}\left(1-z_{nl}u_{ld}\right)\right]\;. \label{eq:indicator}
\end{align}
This allows us to rewrite the likelihood as $\sigma(\lambda)^P \sigma(-\lambda)^{ND-P}$
which can be subsequently maximised with respect to $\lambda$ to yield the update
\begin{align}
  \sigma(\hat{\lambda}) =\frac{P}{ND}\;. \label{eq:lbda_mle}
\end{align}
The alternation between sampling $(\bm{U},\bm{Z})$ and updating the dispersion parameter is carried out until convergence; see Algorithm~\ref{alg:sampler} for all steps of this procedure.

\begin{algorithm}[tb]
  \caption{Sampling from the OrMachine}
  \label{alg:sampler}
  \begin{algorithmic}
    \FOR{$i\;\text{in}\;1,\ldots,\text{max-iters}$}
    \FOR{$n\;\text{in}\;1,\ldots,\text{N}$ \textit{(in parallel)}}
    \FOR{$l\;\text{in}\;1,\ldots,\text{L}$}
    \STATE Compute $p(z_{nl}|\wc)$ following Algorithm \ref{alg:update_z_2}
    \STATE Flip $z_{nl}$ with probability $[p(z_{nl}|\wc)^{-1}{-}1]^{-1}$
    \ENDFOR
    \ENDFOR

    \FOR{$d\;\text{in}\;1,\ldots,\text{d}$ \textit{(in parallel)}}
    \FOR{$l\;\text{in}\;1,\ldots,\text{L}$}
    \STATE Compute $p(u_{ld}|\wc)$ following Algorithm \ref{alg:update_z_2}
    \STATE Flip $u_{ld}$ with probability $[p(u_{ld}|\wc)^{-1}{-}1]^{-1}$
    \ENDFOR
    \ENDFOR
    \STATE Set $\lambda$ to its MLE according to Eq.~\eqref{eq:lbda_mle}.
    \ENDFOR
  \end{algorithmic}
\end{algorithm}

\subsection{Dealing with Missing Data} \label{sec:dealing-with-missing-1}

We can handle unobserved data, by marginalising the likelihood over the missing observations. 
More precisely, if $\bm{X} = (\bm{X}_{\text{obs}}, \bm{X}_{\text{mis}})$ is the decomposition of the full matrix
into the observed part $\bm{X}_{\text{obs}}$ and the missing part $\bm{X}_{\text{mis}}$, after marginalisation, 
the initial likelihood $p(\bm{X}|\bm{U},\bm{Z},\lambda)$ simplifies to 
$p(\bm{X}_{obs}|\bm{U},\bm{Z},\lambda)$. Then, a na\"{\i}ve implementation could be based on indexing 
the observed components inside matrix $\bm{X}$ and modifying the inference procedure so that
the posterior conditionals of $z_{nl}$ and $u_{ld}$ involve only sums over observed elements. A simpler, equivalent implementation, which we follow in our experiments, is to      
represent the data as $\tilde{x}_{nd} \in \{-1,0,1\}$ where missing observations are encoded as zeros, each contributing the constant factor $\sigma(0)=\nicefrac{1}{2}$ to the full
likelihood, so that 
\begin{align}
  p(\bm{X}|\bm{U},\bm{Z},\lambda) = C\, p(\bm{X}_{obs}|\bm{U},\bm{Z},\lambda)\;,
\end{align}
where $C$ is a constant. 
Thus, the missing values do not contribute to the posterior over $\bm{U}$ and $\bm{Z}$ which is also 
clear from the form of the full conditionals in Eq.~\eqref{eq:update_z} that depend on a sum weighted by 
$x_{nd}$s. For the update of the dispersion parameter in Eq.~\eqref{eq:lbda_mle}), we need to subtract the number of all missing observations in the denominator. The dispersion now indicates the fraction 
of correct prediction in the observed data.
Following this inference procedure, we can impute missing data based on a Monte Carlo estimate of the predictive 
distribution of some unobserved $x_{nd}$ as
\begin{align}
\frac{1}{S} \sum_{s=1}^{S} p(x_{nd}|\bm{U}^{(s)},\bm{Z}^{(s)}, \hat{\lambda})\;, 
\end{align}
where each $(\bm{U}^{(s)},\bm{Z}^{(s)})$ is a posterior sample. A much faster approximation of the predictive distribution is obtained by $p(x_{ns}|\hat{\bm{U}},\hat{\bm{Z}}, \hat{\lambda})$, where we simply plug the posterior mean estimates for $(\bm{U}, \bm{Z})$ into the predictive distribution. 
For the simulated data in Section~\ref{sec:completion}, we find both methods to perform equally well and therefore follow the second, faster approach for all remaining experiments.

\subsection{Multi-Layer OrMachine}
BooMF learns patterns of correlation in the data. In analogy to multi-layer neural networks, we can build a hierarchy of correlations by applying another layer of factorisation to the factor matrix $\bm{Z}$. This is reminiscent of the idea of deep exponential families, as introduced by \citet{ranganath2015deep}.
The ability to learn features at different levels of abstraction is commonly cited as an explanation for the success that deep neural networks have across many domains of application~\cite{lin2016_why, bengio13:_repres_learn}. In the present setting, with stochasticity at every step of the generative process and posterior inference, we are able to infer meaningful and interpretable hierarchies of abstraction.
\begin{figure}[tb]
  \centering
  \includegraphics[width=.95\linewidth]{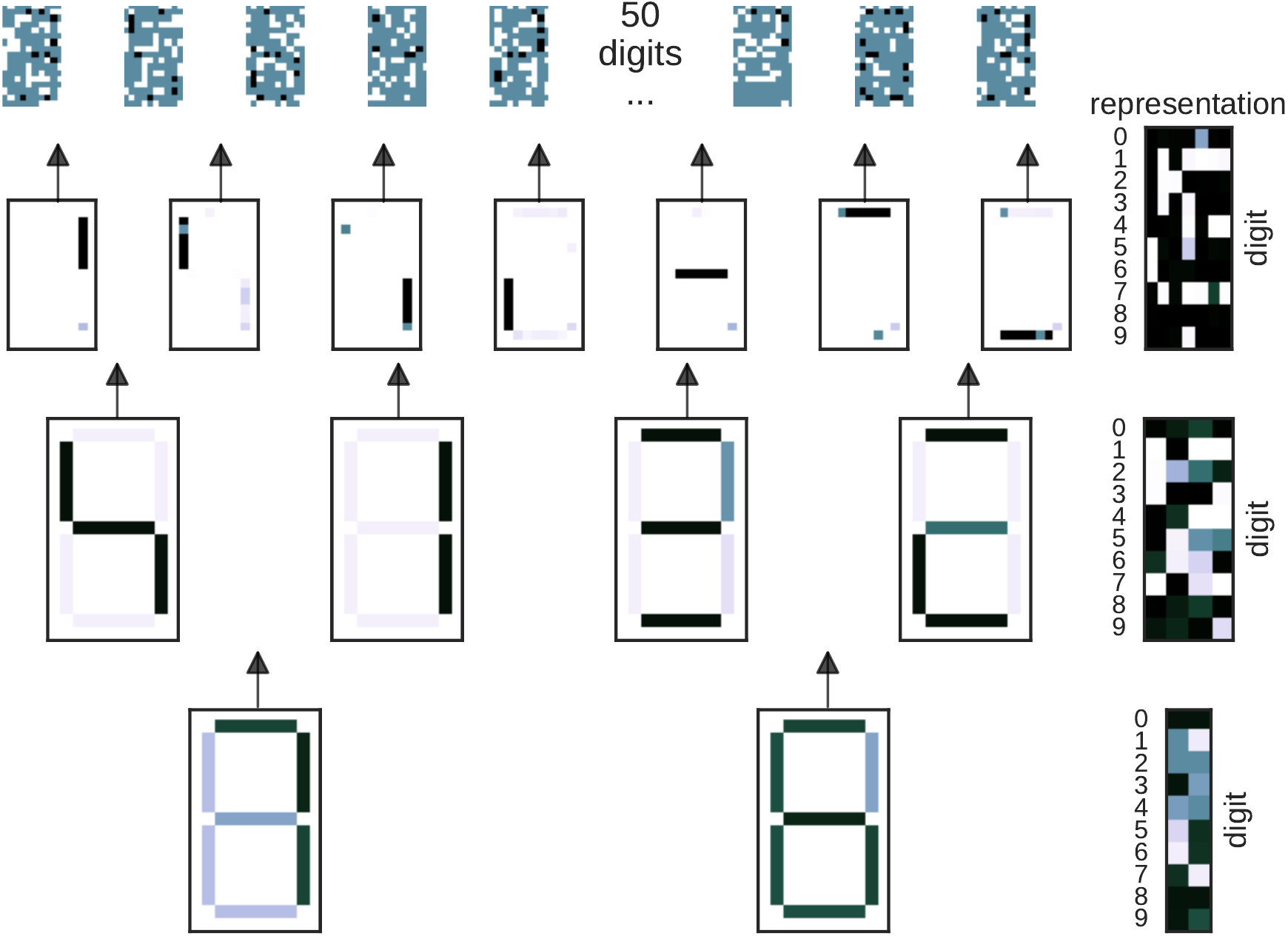}
  \caption{An OrMachine with 3 hidden layers is trained to reconstruct 50 calculator digits with 70\% of observations missing. The rows depict increasingly abstract layers of the model. Shown are the latent prototypes fed forward to the data layer. Variables are arranged to $17{\times}10$ images for interpretation. The right sides show the corresponding posterior means for representations of the partially observed input digits.}
  \label{fig:deep_calc}
\end{figure}
\begin{figure*}[tb]
  \centering
    \includegraphics[width=.155\linewidth]{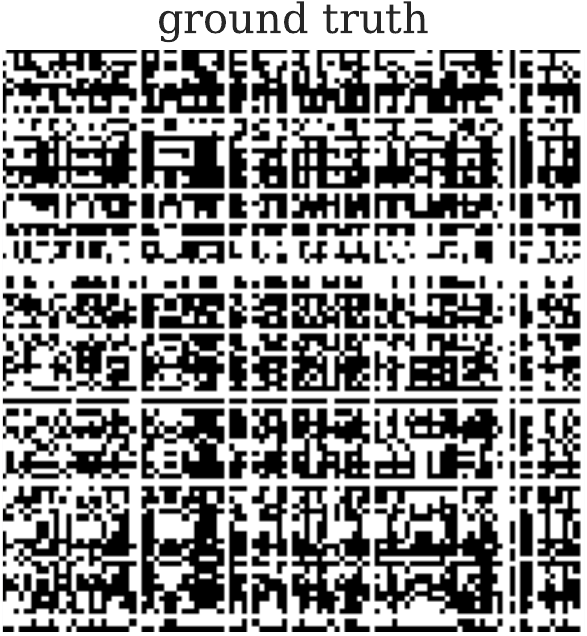}\hspace{.05\linewidth}
   \includegraphics[width=.155\linewidth]{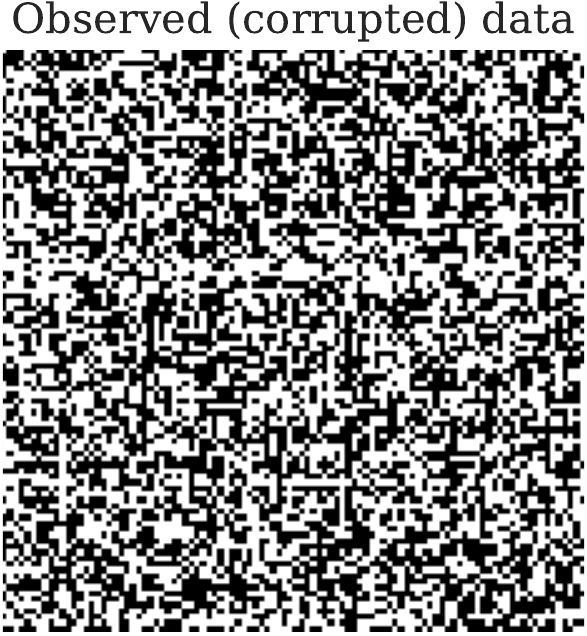}\hspace{.05\linewidth}
   \includegraphics[width=.21\linewidth]{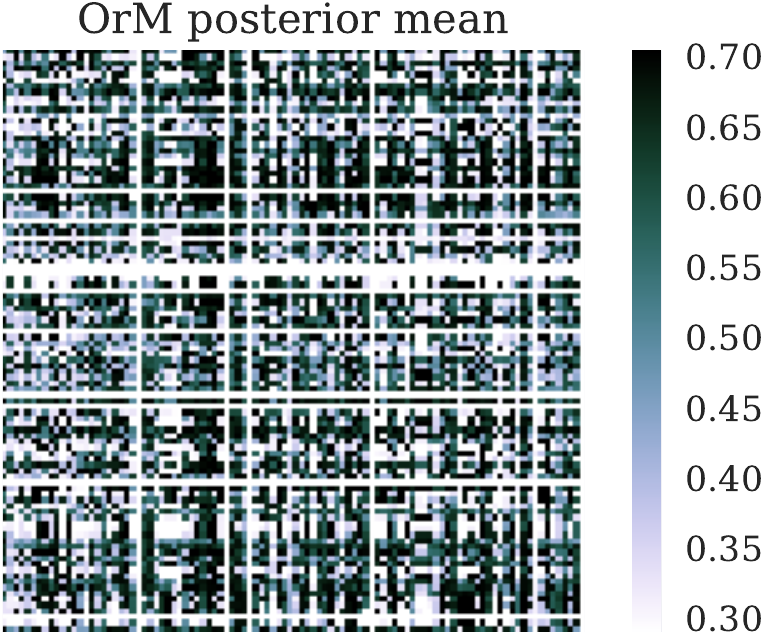}\hspace{.05\linewidth}
   \includegraphics[width=.16\linewidth]{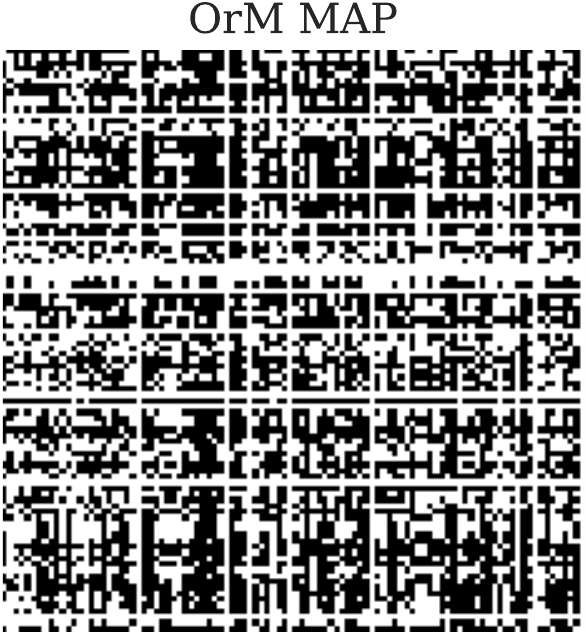}
  \caption{Illustration of the matrix factorisation task for a $100 \times 100$ matrix of rank 7. The posterior means estimate the probability of each data point to take a value of one. MAP estimates are computed by rounding to the closest integer.}
  \label{fig:recon_example}
\end{figure*}

To give an example, we determine the optimal multi-layer architecture for representing the calculator digit toy dataset as introduced in Fig.~\ref{fig:calc}. We observe 50 digits and consider $70\%$ of the data points randomly as unobserved. We then train multi-layer OrMachines with various depths and layer widths, iterating through the individual layers during 200 iterations of burn-in. We then we draw 200 samples from each consecutive layer with the remaining layers held fixed to their MAP estimate. In order to enforce distributed representations, we choose independent Bernoulli sparsity priors for the codes: $p(u_{ld})=[0.01, 0.05, 0.2]$ for each layer, respectively. Superior performance in reconstructing the unobserved data is achieved by a 3-hidden layer architecture with hidden layers of size $L_1=7$, $L_2=4$, $L_3=2$. This 3-layer model reduces the reconstruction error from $1.4\%$ to $0.4\%$ compared to the single-layer model with width $L=7$. Maximum likelihood estimates of the dispersion for the three layers are $\hat{\bm{\lambda}}=[1.0, 0.93, 0.8]$.
The first layer infers the seven bars that compose all digits. We plot the probabilities that each prototype induces in the observation layer, given by the one-hot activations of $\bm{z}_{l=1\ldots L}$ in Fig.~\ref{fig:deep_calc}. They are depicted alongside the average posterior mean of the representations for each digit in the training data. This example illustrates that the multi-layer OrMachine infers interpretable higher-order correlations and is able to exploit them to achieve significant improvements in missing data imputation.

\subsection{Practical Implementation and Speed}
The algorithm is implemented in Python with the core sampling routines in compiled Cython. The binary data is represented as $\{-1,1\}$ with missing data encoded as $0$. This economical representation of data and variables as integer types simplifies computations considerably. Algorithm~\ref{alg:update_z_2} is implemented in parallel across the observations $[n]=\{1,\dots, N\}$ and conversely updates for $u_{ld}$ are implemented in parallel across all features $[d]=\{1,\dots, D\}$.

The computation time scales linearly in each dimension. A single sweep through high-resolution calculator digits toy dataset with $ND=1.7\times 10^6$ data points and $L=7$ latent dimensions takes approximately 1 second on a desktop computer. A single sweep through the approximately $1.4\times10^{10}$ data points presented in the biological example in Section~\ref{sec:expl-analys-single} with $L=2$ latent dimensions takes approximately 5 minutes executed on 24 computing cores. For all examples presented here 10--20 iterations suffice for the algorithm to converge to a (local) posterior mode.

\section{Experiments on Simulated Data}
\label{sec:exper-simul-data}
In this section, we probe the performance of the OrMachine (OrM) at random matrix factorisation and completion tasks.
Message passing (MP) has been shown to compare favourably with other state-of-the-art methods for BooMF~\cite{ravanbakhsh2015_boolean-matrix} and is therefore the focus of our comparison. The following settings for MP and the OrM are used throughout our experiments, unless mentioned otherwise. For MP, we use the Python implementation provided by the authors. We also proceed with their choice of hyper-parameters, as experimentation with different learning rates and maximum number of iterations did not lead to any improvements.
For both methods, we set the priors $p(u)$ and $p(z)$ to the factor matrices' expected value based on the density of the product matrix in an Empirical-Bayes fashion. The only exception is MP in the matrix completion task, where uniform priors, as used by \citet{ravanbakhsh2015_boolean-matrix}, lead to slightly better performance. For the OrM, we initialise the parameters uniformly at random and draw 100 iterations after 100 samples of burn-in. Note that around 10 sampling steps are usually sufficient for convergence.

\subsection{Random Matrix Factorisation}
\label{sec:factorisation}
We generate a quadratic matrix $\bm{X}\in\{0,1\}^{N\times N}$ of rank $L$ by taking the Boolean product of two random $N\times L$ factor matrices. The Boolean product $X$ of two rank $L$ binary matrices that are sampled i.i.d.\ from a Bernoulli distribution with parameter $p$ has an expected value of \mbox{$E(X)=1-(1-p^2)^L$}.
Since we generally prefer $X$ to be neither sparse nor dense, we fix its expected density to $\nicefrac{1}{2}$, unless stated otherwise. This ensures that a simple bias toward zeroes or ones in either method is not met with reward.
\begin{figure}[tb]
  \centering
  \belowspace
  \includegraphics[width=.89\linewidth]{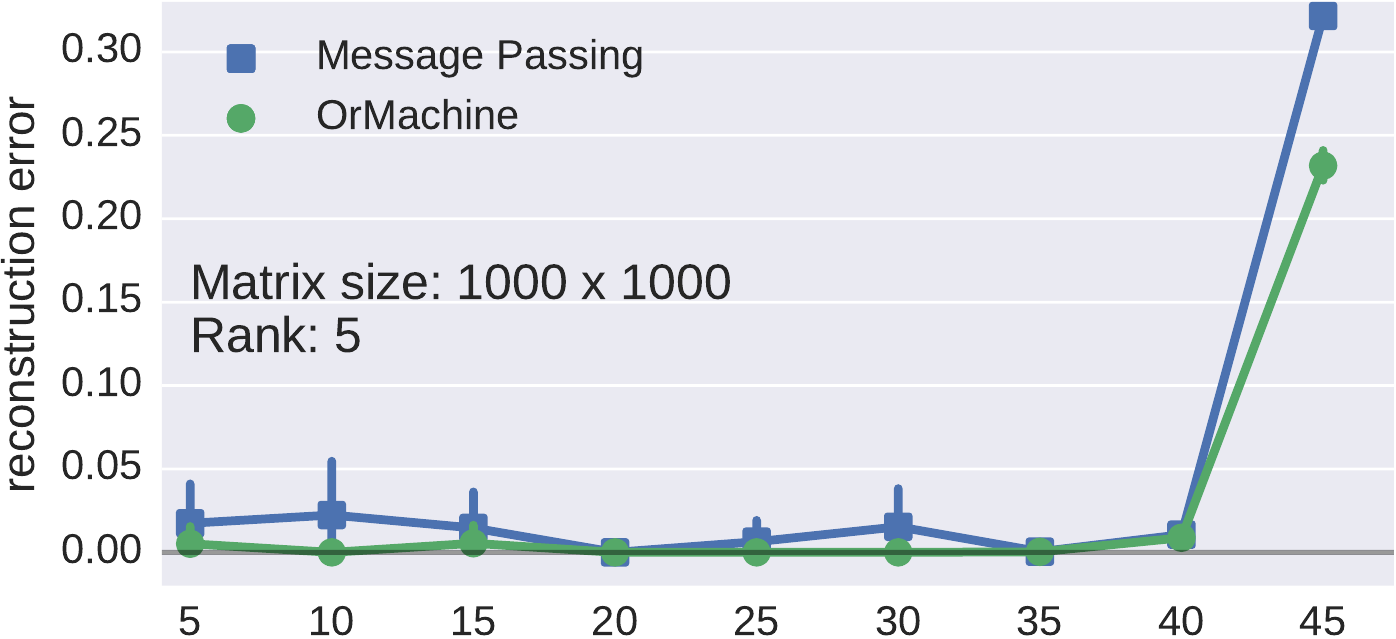} 
  \belowspace
  \includegraphics[width=.89\linewidth]{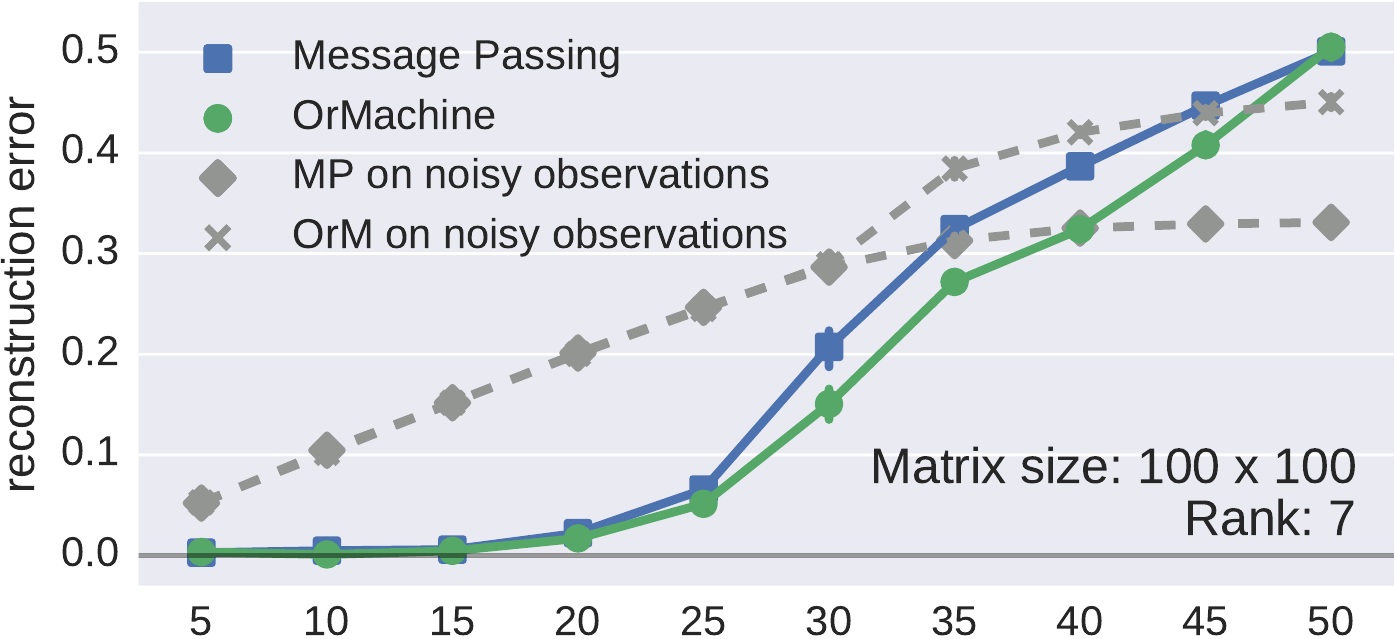} 
  \includegraphics[width=.89\linewidth]{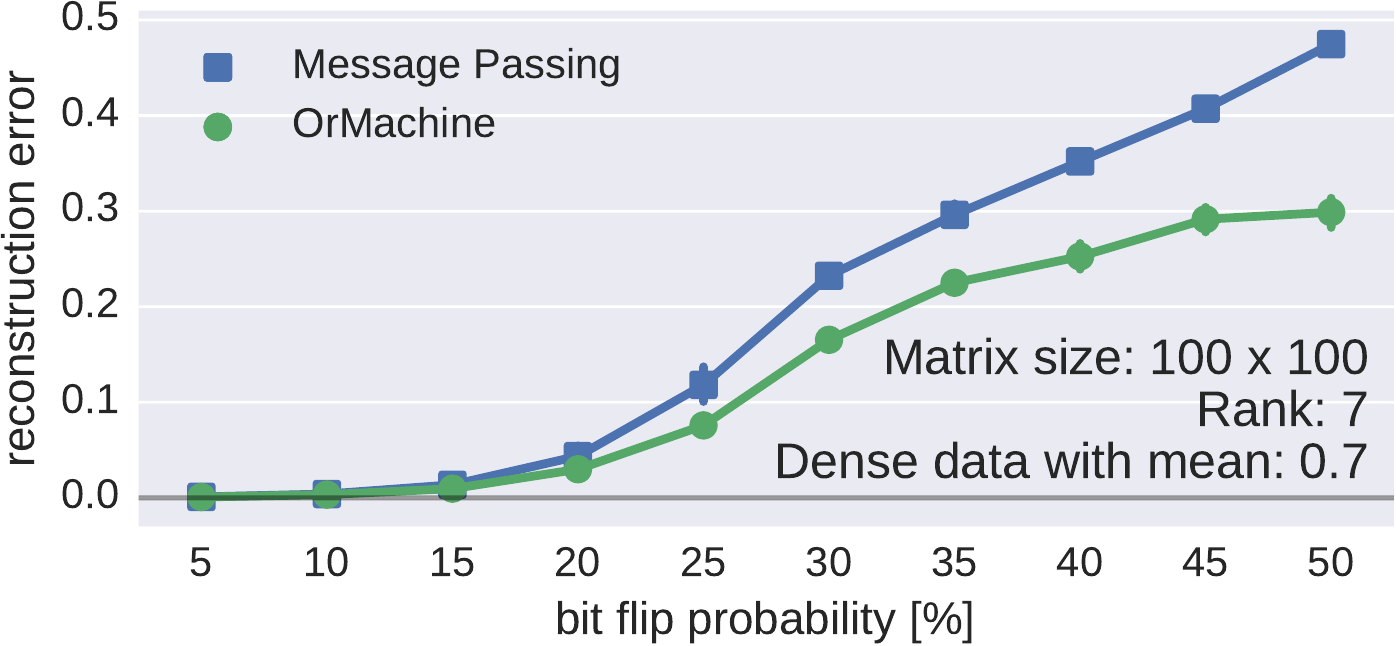}
  \caption{Comparison of OrMachine and message passing for BooMF for random matrices of different size, rank and density. Compare to Fig.~2 in \citet{ravanbakhsh2015_boolean-matrix}.}
  \label{fig:recon_cmprs}
\end{figure}
Bits in the data are flipped at random with probabilities ranging from $5\%$ to $50\%$. Factor matrices of the correct underlying dimension are inferred and the data is reconstructed from the inferred factorisation.
An example of the task is shown in Fig.~\ref{fig:recon_example}.

Results for the reconstruction error, defined as the fraction of correctly reconstructed data points, are depicted in Fig.~\ref{fig:recon_cmprs}. All experiments were repeated 10 times with error bars denoting standard deviations.
The OrM outperforms MP under all conditions, except when both methods infer equally error-free reconstructions. Fig.~\ref{fig:recon_cmprs}~(top) reproduces the experimental settings of Fig.~2 in \citet{ravanbakhsh2015_boolean-matrix}.
We find that the OrMachine enables virtually perfect reconstruction of a $1000\times1000$ matrix of rank $L=5$ for up to $35\%$ bit flip probability. Notably, MP performs worse for smaller noise levels. It was hypothesised by \citet{ravanbakhsh2015_boolean-matrix} that symmetry breaking at higher noise levels helps message passage to converge to a better solution.
Fig.~\ref{fig:recon_cmprs}~(middle) demonstrates the consistently improved performance of the OrMachine for a more challenging example of $100\times100$ matrices of rank 7. The reconstruction performance of both methods is similar for lower noise levels, while the OrMachine consistently outperforms MP for larger noise levels.
For biased data with $\mathbb{E}[x_{nd}]=0.7$ in Fig.~\ref{fig:recon_cmprs}~(bottom), we observe a similar pattern with a larger performance gap for higher noise levels. Even for a bit flip-probability of $50\%$ the OrMachine retains a reconstruction error of approximately $30\%$, which is achieved by levering the bias in the data.

Fig.~\ref{fig:recon_cmprs} (middle) also shows the reconstruction error on the observed data, indicating that MP overfits the data more than the OrM for larger noise levels. This may contribute to the improved performance of the OrMachine.


\subsection{Random Matrix Completion}
\label{sec:completion}

\begin{figure}[tb]
  \centering
  \includegraphics[width=\linewidth]{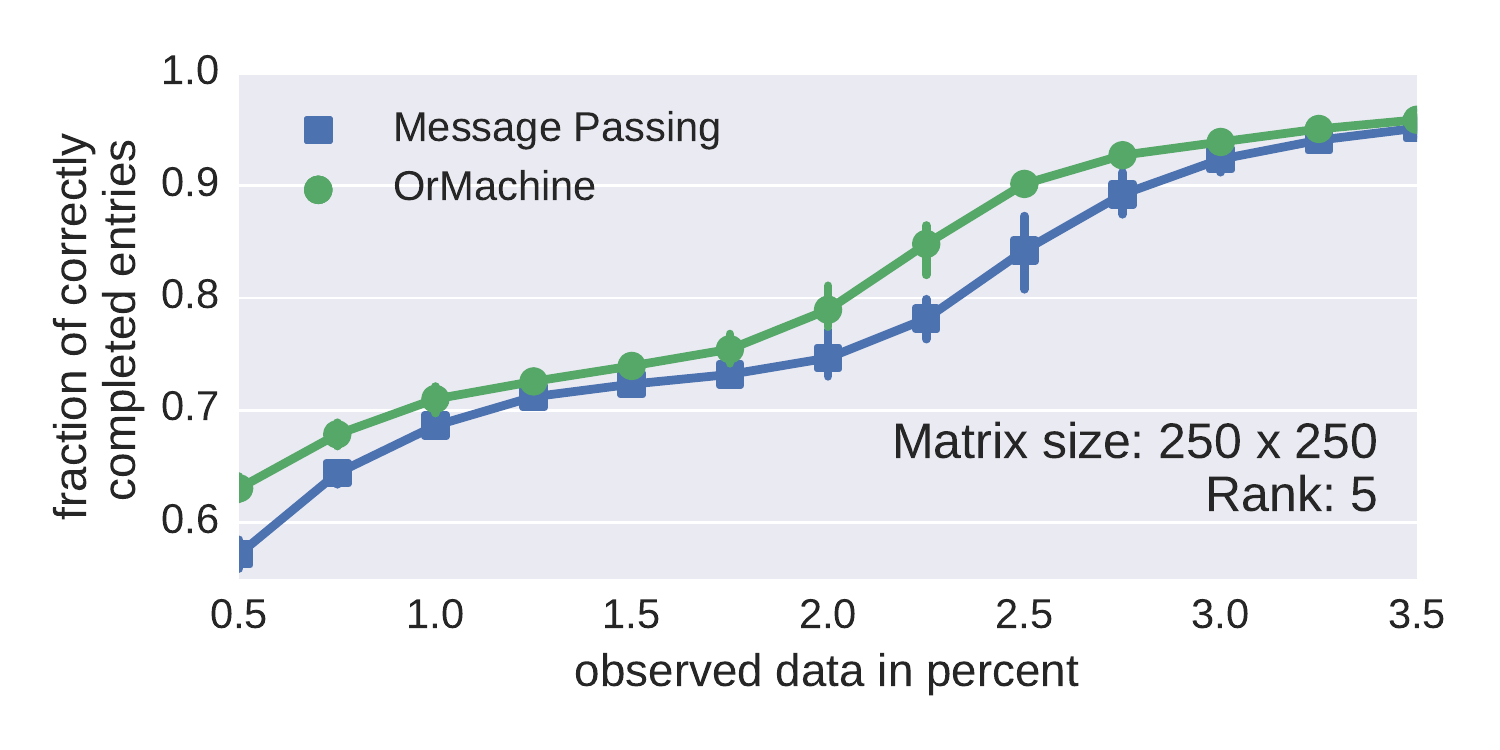}
  \includegraphics[width=\linewidth]{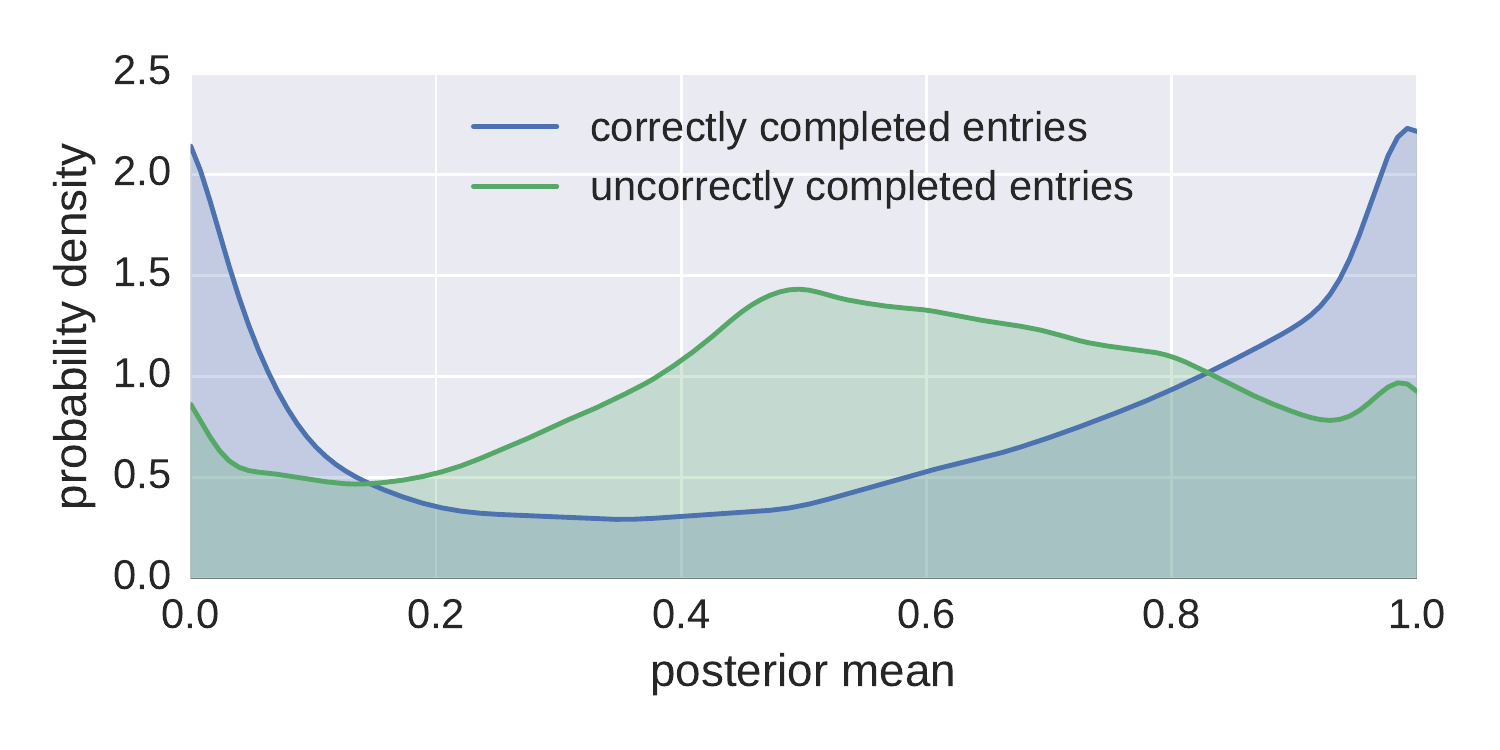}
  \caption{Matrix completion performance for simulated low rank matrices (top) and kernel density estimate of the distribution of posterior means for inferred matrix entries (bottom).}
  \label{fig:cmprs}
\end{figure}

We further investigate the problem of matrix completion or collaborative filtering, where bits of the data matrix are unobserved and reconstructed from the inferred factor matrices. Following the procedure outlined in Section~\ref{sec:factorisation}, we generate random matrices of rank $5$ and size $250\times250$. We only observe a random subset of the data, ranging from $0.5\%$ and $3.5\%$. The missing data is reconstructed from the inferred factor matrices. As shown in Fig.~\ref{fig:cmprs}, the OrMachine outperforms message passing throughout. The plot indicates means and standard deviations from 10 repetitions of each experiment.

Notably, the OrMachine does not only provide a MAP estimate, but also an estimate of the posterior probability for each unobserved data point $x_{nd}$. Fig.~\ref{fig:cmprs}~(bottom) shows an estimate of the density of the posterior means for the correctly and incorrectly completed data points. The distribution of incorrect predictions peaks around a probability of $\nicefrac{1}{2}$, indicating that the OrMachine's uncertainty about its reconstruction provides further useful information about the missing data. For instance, this information can be used to control for false positives or false negatives, simply by setting a threshold for the posterior mean.

\section{Experiments on Real-World Data}
\label{sec:exper-real-world}

\subsection{MovieLens Matrix Completion}

\begin{table}[tb]
  \caption{Collaborative filtering performance for MovieLens 1M and 100k dataset. Given are the percentages of correctly reconstructed unobserved data as means from 10 random repetitions. Compare to Table 1 in \citet{ravanbakhsh2015_boolean-matrix}, who also provide comparison to other state-of the art methods. Their results for message passing were independently reproduced. The multi-layer OrMachine has two hidden layers of size 4 and 2, respectively.} 
\label{tab:movie_lense}
\vskip 0.1in
\begin{center}
\begin{footnotesize}
\begin{sc}
  \begin{tabular}{lcccccc}
    & \multicolumn{6}{c}{observed percent. of available ratings} \\
    &1\%&5\%&10\%&20\%&50\%&95\%\\
    \hline
    \abovespace
    \textbf{100K}& & & & & & \\
    OrM &58.5&63.5&64.9&66.4&68.9&70.0\\    
    \belowspace
    MP &52.8&60.7&63.0&65.2&67.5&69.5\\
    \begin{minipage}{.01\textwidth} \tiny{Multi\\ \;layer} \end{minipage} & \multirow{2}{*}{58.5} & \multirow{2}{*}{63.5}
         & \multirow{2}{*}{65.2} & \multirow{2}{*}{66.5}
         & \multirow{2}{*}{68.8} & \multirow{2}{*}{70.1} \\
    \;OrM &&&&&& \\
    \abovespace
    \textbf{1M}& & & & & & \\
    OrM &63.4&67.0&68.5&69.8&70.9&71.2\\    
    \belowspace
    MP &56.7&64.9&67.2&68.8&70.7&71.5\\
    \begin{minipage}{.01\textwidth} \tiny{Multi\\ \;layer} \end{minipage} & \multirow{2}{*}{63.8} & \multirow{2}{*}{67.2}
         & \multirow{2}{*}{68.6} & \multirow{2}{*}{70.0}
                 & \multirow{2}{*}{71.4} & \multirow{2}{*}{72.1} \\
    \belowspace
    \;OrM &&&&&& \\

\hline
\end{tabular}
\end{sc}
\end{footnotesize}
\end{center}
\vskip -0.1in
\end{table}
\begin{figure}[tb]
  \centering
  \includegraphics[width=.98\linewidth]{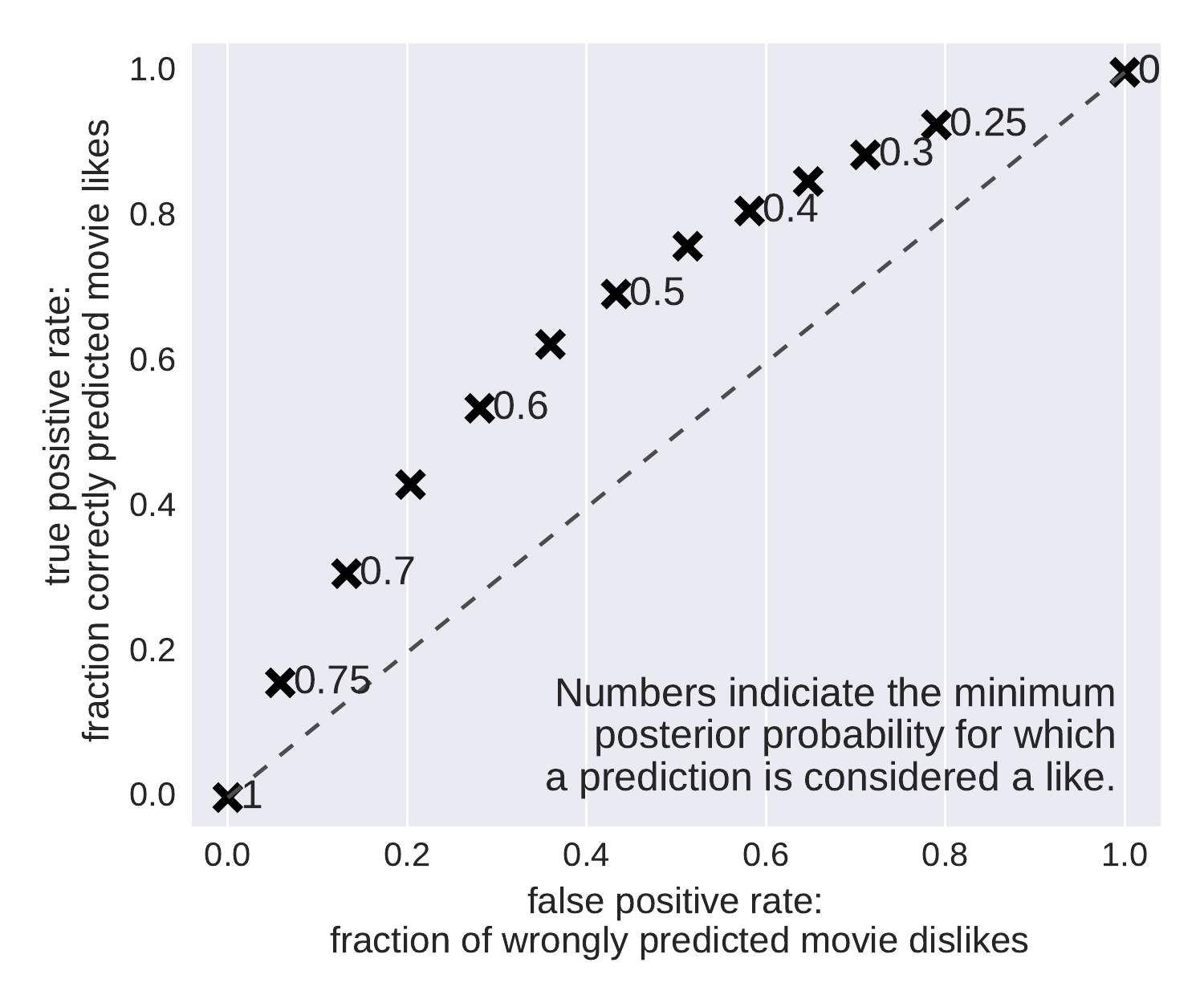}
  \caption{ROC curve for MovieLens 100k data, adjusting the threshold for when a prediction is considered a like. 10\% of the available data were observed and used for inference with an OrM of size $L=2$. Predictions were tested on the remaining 90\%.}
  \label{fig:recon_roc}
\end{figure}

We investigate the OrMachine's performance for collaborative filtering on a real-world dataset. The MovieLens-1M dataset\footnote{The MovieLens dataset is available online: \url{https://grouplens.org/datasets/movielens/}.} contains $10^6$ integer film ratings from 1 to 5 from 6000 users for 4000 films, i.e.\ 1/24 of the possible ratings are available. Similarly, the MovieLens 100k dataset contains 943 users and 1682 films. Following \citet{ravanbakhsh2015_boolean-matrix}, we binarise the data taking the global mean as threshold. We observe only a fraction of the available data, varying from $1\%$ to $95\%$, and reconstruct the remaining available data following the procedure in Section~\ref{sec:completion} with $L=2$ latent dimensions. Reconstruction accuracies are given as fractions of correctly reconstructed unobserved ratings in Table~\ref{tab:movie_lense}.
The given values are means from 10 randomly initialised runs of each algorithm. The corresponding standard deviations are always smaller than $0.2\%$. The OrMachine is more accurate than message passing in all cases, except for the 1M dataset with 95\% available ratings. The OrMachine's advantage is particularly significant if only little data is observed.
Increasing the latent dimension $L$ to values of $3$ or $4$ yields no consistent improvement, while a further increase is met with diminishing returns.
We achieve the best within-sample performance for a two-layer OrMachine with different architectures performing best for different amounts of observed data. An OrMachine with two hidden layers of sizes 4 and 2 respectively yields the best average performance. As indicated in Table~\ref{tab:movie_lense}, it provides better results throughout but exceeds the performance of the shallow OrMachine rarely by more than 1\%. This indicates that there is not much higher order structure in the data, which is unsurprising given the sparsity of the observations and the low dimensionality of the first hidden layer.

We illustrate a further advantage of full posterior inference for collaborative filtering. We can choose a threshold for how likely we want a certain prediction to take a certain value and trade off false with true positives. A corresponding ROC curve for the MovieLens 100k dataset, where 10\% of the available data was observed, is shown in Fig.~\ref{fig:recon_roc}.

\subsection{Explorative Analysis of Single Cell Gene Expression Profiles}
\label{sec:expl-analys-single}
Single-cell RNA expression analysis is a revolutionary experimental technique that facilitates the measurement of gene expression on the level of a single cell~\cite{Blainey2014}. In recent years this has led to the discovery of new cell types and to a better understanding of tissue heterogeneity~\cite{trapnell2015_defin}. The latter is particularly relevant in cancer research where it helps to understand the cellular composition of a tumour and its relationship to disease progression and treatment~\cite{patel2014_singl-rna}.
Here we apply the OrMachine to binarised gene expression profiles of about 1.3 million cells for about 28 thousand genes per cell. Cell specimens were obtained from cortex, hippocampus and subventricular zone of E18 (embryonic day 18) mice; the data is publicly available\footnote{\url{https://support.10xgenomics.com}}.
Only 7\% of the data points are non-zero. We set all non-zero expression levels to one, retaining the essential information of whether or not a particular gene is expressed.
We remove genes that are expressed in fewer than 1\% of cells with roughly 11 thousand genes remaining. This leaves us with approximately $1.4\times 10^{10}$ data points. We apply the OrMachine for latent dimensions $L=2,\ldots,10$. The algorithm converges to a posterior mode after 10--20 iteration, taking roughly an hour on a 4-core desktop computer and 10--30 minutes on a cluster with 24 cores. We draw 125 samples and discard the first 25 as burn-in.
\begin{figure}[tb]
  \centering
  \includegraphics[width=\linewidth]{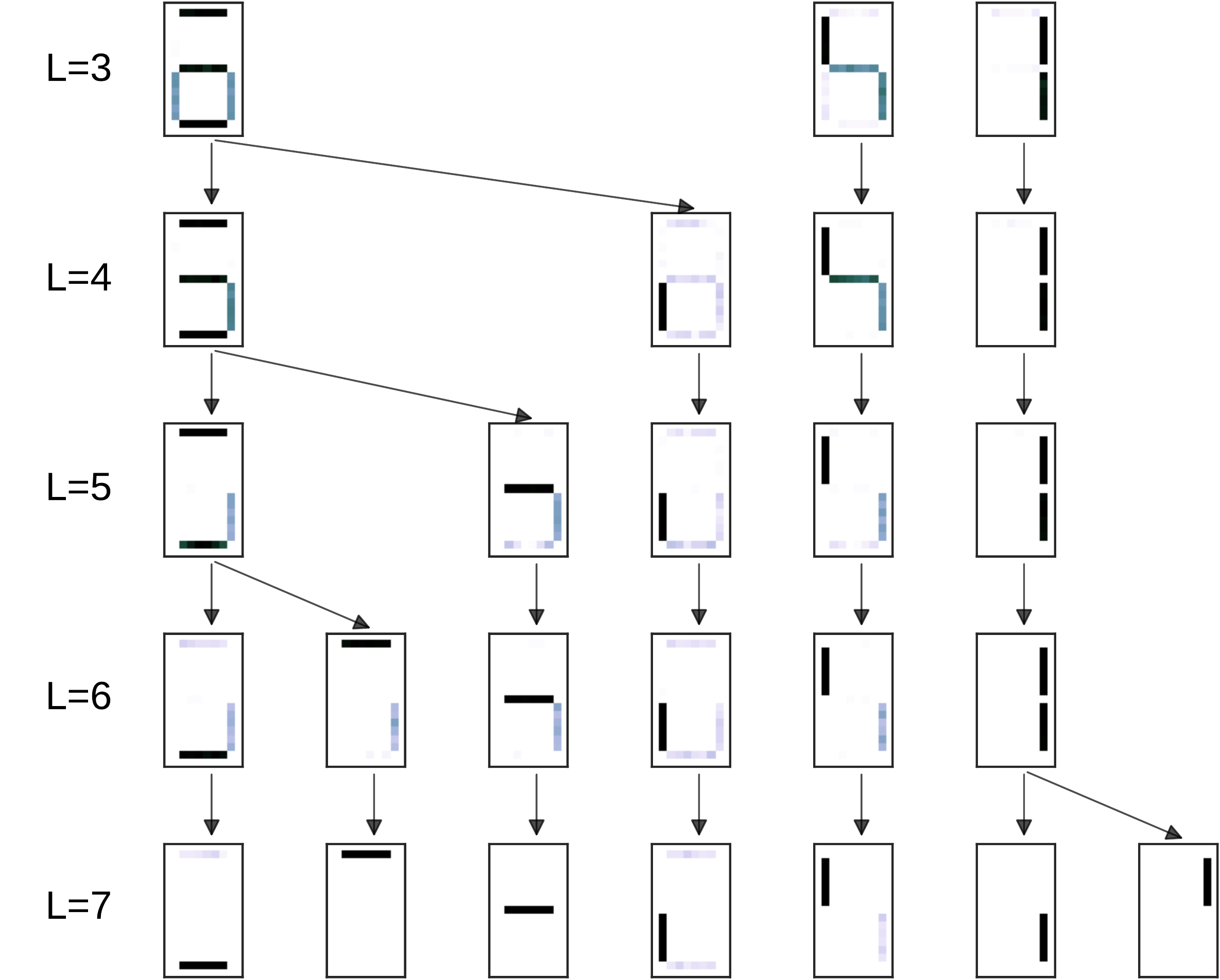}
  \caption{Hierarchy in the latent features of calculator digits. The arrows indicate the split of codes into more distributed codes of lower density. They can be inferred as latent variables in an OrMachine, with codes from the model of size $L$ as data and codes from model with of size $L{+}1$ as fixed codes. \label{fig:calc_hier}}
\end{figure}

\begin{figure*}[tb]
  \centering  \tcbox[sharp corners, boxsep=-1mm,colback=white,boxrule=0.2mm]{\includegraphics[width=.965\linewidth]{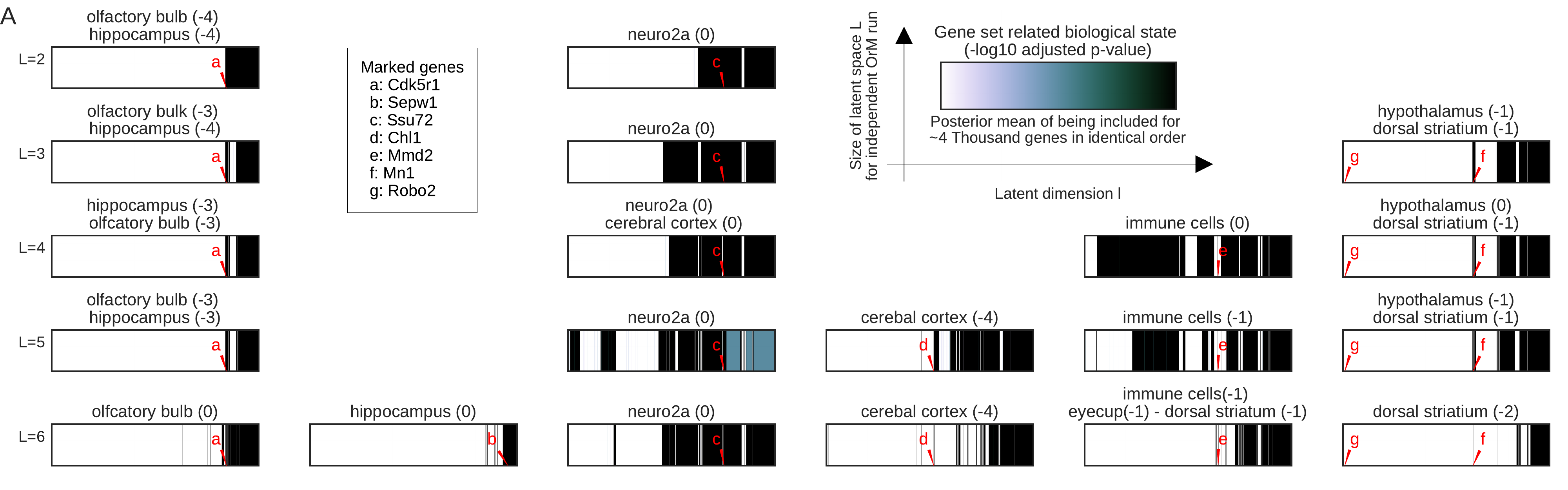}}
  \tcbox[sharp corners, boxsep=-1mm,colback=white,boxrule=0.2mm]{\includegraphics[width=.965\linewidth]{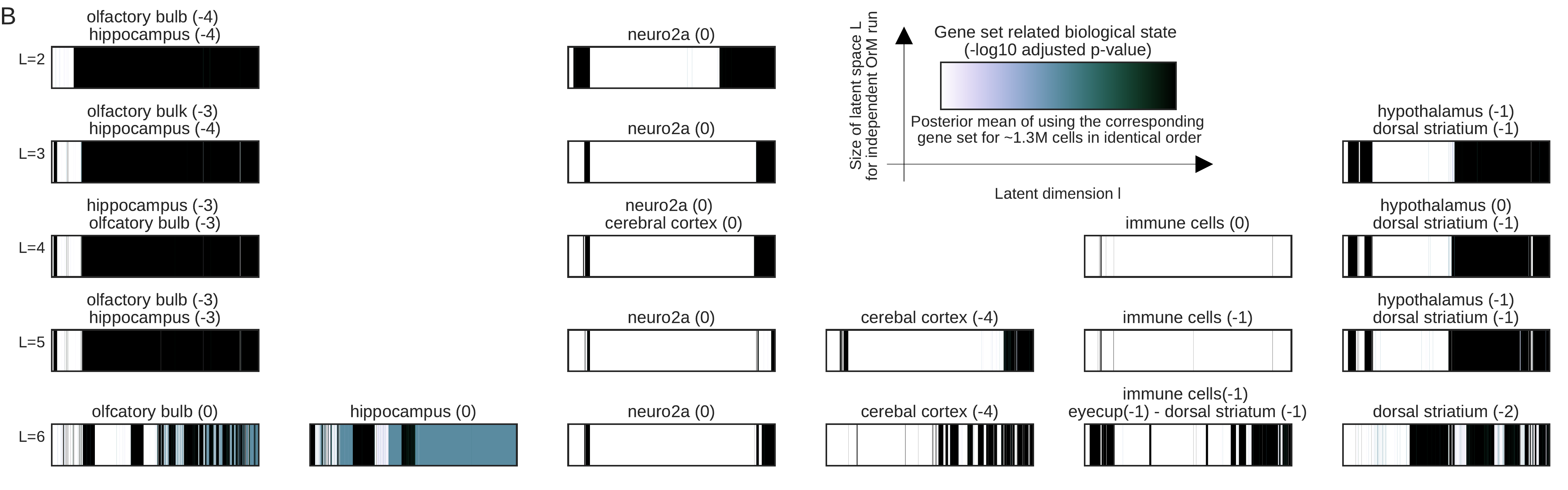}}
  \caption{Hierarchy in the latent representations of genes (A) and specimens (B) under variation of the latent dimensionality. Rows with the same dimensionality ($L=2\ldots6$) in the top and bottom box correspond to the same OrMachine factorisation. Rows with different dimensionality are trained independently. Importantly, the ordering of genes/specimens is identical for all subplots. Each subplot (A) describes set of expressed genes, limited to the approximately 4k out of 11k analysed genes that are used in at least one code. Subplots in (B) describe representations of cell specimens in terms of which of the gene sets they express. See legend in each box for more details.}
\label{fig:neurons}
\end{figure*}
Factorisations with different latent dimensionality form hierarchies of representations, where features that appear together in codes for lower dimensions are progressively split apart when moving to a higher dimensional latent space. We illustrate our approach to analysing the inferred factorisations on calculator digits in Fig.~\ref{fig:calc_hier}. Each row corresponds to an independently trained OrMachine with the $L$ increasing from 3 to 7. We observe denser patterns dividing up consecutively until only the seven constituent bars remain. This is a form of hierarchical clustering that, in contrast to traditional methods, does not impose any hierarchical structure on the model.
We perform the same analysis on the single cell gene expression data with the results for both, gene patterns and specimen patterns shown in Fig.~\ref{fig:neurons}. Furthermore, we run a gene set enrichment analysis for the genes that are unique to each inferred code, looking for associated biological states. This is done using the Enrichr analysis tool~\cite{chen2013_enric} and a mouse gene atlas~\cite{su2004}. Biological states are denoted together with the logarithm to base 10 of their adjusted p-value.
Increasing the latent dimensionality leads to a more distributed representation with subtler, biologically plausible patterns. The columns in Fig.~\ref{fig:neurons} are ordered to emphasise the hierarchical structure within the gene sets and their assignments. For example, in the first column for $L=5$ and second column for $L=6$, a gene set with significant overlap to two biological processes (olfactory bulb and hippocampus) splits into two gene sets each corresponding to one of the two processes. In the specimen assignments (\ref{fig:neurons}B) this is associated with an increase in posterior uncertainty as to which cell expresses this property. The significance levels of the associated biological processes drop from p-values on the order of $10^{-3}$ to p-values on the order of $1$. Typical genes for each of the biological states are annotated \cite{lopez-bendito2007_robo1-robo2,zheng2008_gene-huang,demyanenko2010_l1-chl1,upadhya2011_expres,raman2013_selen-w}.
This examples illustrates the OrMachine's ability to scale posterior inference to massive datasets. It enables the discovery of readily interpretable patterns, representations and hierarchies, all of which are biologically plausible.

\section{Conclusion}
\label{sec:conclusion}
We have developed the OrMachine, a probabilistic model for Boolean matrix factorisation. The extremely efficient Metropolised Gibbs sampler outperforms state-of-the-art methods in matrix factorisation and completion. It is the first method that infers posterior distributions for Boolean matrix factorisation, a property which is highly relevant in practical applications where full uncertainty quantification matters.
Despite full posterior inference, the proposed method scales to very large datasets. We have shown that tens of billions of data points can be handled on commodity hardware. The OrMachine can readily accommodate missing data and prior knowledge.
Layers of OrMachines can be stacked, akin to deep belief networks, inferring representations at different levels of abstraction. This leads to improved reconstruction performance in simulated and real world data.

Future work will include further experiments on the ability to learn deep probabilistic abstractions, as well as more principled methods to infer the optimal model architecture.





\begingroup
\sloppy
\bibliography{MyLibrary}
\bibliographystyle{icml2017}
\endgroup

\end{document}